\documentclass[journal]{IEEEtran}
\usepackage{amsmath}
\usepackage{amssymb}
\usepackage{mathrsfs}
\usepackage{cite} 
\usepackage{booktabs}
\usepackage[table,xcdraw]{xcolor}
\usepackage[normalem]{ulem}
\useunder{\uline}{\ul}{}
\usepackage{multirow}
\usepackage{graphicx}
\usepackage{subfigure} 
\usepackage{caption}
\usepackage[ruled,linesnumbered]{algorithm2e} 
\usepackage{hyperref}
\usepackage{url}
\usepackage{epstopdf}
\usepackage{lineno}
\begin{document}
	
%\switchlinenumbers
%\pagewiselinenumbers

\title{Hard Region Aware Network for Remote Sensing Change Detection}

\author{Zhenglai Li, 
	Chang Tang, ~\IEEEmembership{Senior Member,~IEEE,}
	Xinwang Liu, ~\IEEEmembership{Senior Member,~IEEE,}
	Xingchen Hu, \\
	Xianju Li,
	Ning Li,
    Changdong Li
\thanks{Manuscript received X XX, XXXX; revised X XX, XXXX. This work was supported in part by the National Science Foundation of China under Grant 62076228 and 62176242, and in part by Natural Science Foundation of Shandong Province under Grant ZR2021LZH001. (Corresponding author: Chang Tang.)}

\thanks{Z. Li is with the faculty of data science, City University of Macau, Macau, 00853, China. 
	(E-mail: zlli@cityu.edu.mo).}
\thanks{C. Tang and X. Li are with the school of computer, China University of Geosciences, Wuhan, 430074, China. 
	(E-mail: \{tangchang, ddwhlxj\}@cug.edu.cn).}
\thanks{X. Liu is with the school of computer, National University of Defense Technology, Changsha 410073, China.
	(E-mail:  xinwangliu@nudt.edu.cn).}
\thanks{X. Hu is with the College of Systems Engineering, National University of Defense Technology, Changsha 410073, China. 
	(E-mail: xhu4@ualberta.ca)}
\thanks{N. Li is Key Laboratory of Target Cognition and Application Technology (TCAT), Aerospace Information Research Institute, Beijing, 100190, China.
	(E-mail: lining2434@gmail.com).}
\thanks{C. Li is with the faculty of engineering, China University of Geosciences, Wuhan, 430074, China. 
	(E-mail: lichangdong@cug.edu.cn).}
}

\markboth{Journal of \LaTeX\ Class Files,~Vol.~14, No.~8, August~2015}%
{Shell \MakeLowercase{\textit{et al.}}: Bare Demo of IEEEtran.cls for IEEE Journals}



\maketitle

\begin{abstract}
Change detection (CD) is essential for various real-world applications, such as urban management and disaster assessment. Numerous CD methods have been proposed, and considerable results have been achieved recently. However, detecting changes in hard regions, i.e., the change boundary and irrelevant pseudo changes caused by background clutters, remains difficult for these methods, since they pose equal attention for all regions in bi-temporal images. This paper proposes a novel change detection network, termed as HRANet, which provides accurate change maps via hard region mining. Specifically, an online hard region estimation branch is constructed to model the pixel-wise hard samples, supervised by the error between predicted change maps and corresponding ground truth during the training process. A cross-layer knowledge review module is introduced to distill temporal change information from low-level to high-level features, thereby enhancing the feature representation capabilities. Finally, the hard region aware features extracted from the online hard region estimation branch and multi-level temporal difference features are aggregated into an unified feature representation to improve the accuracy of CD. Experimental results on two benchmark datasets demonstrate the superior performance of HRANet in the CD task.
\end{abstract}


\IEEEpeerreviewmaketitle

\section{Introduction}
Change detection (CD) aims to locate and segment the changes that are most relevant in the bi-temporal images captured at diverse periods in the same surface regions. As an important technique in remote sensing sense understanding, CD has received increased attention and been applied for many real-world applications, such as land-use change detecting~\cite{deng2008pca, hu2018automatic}, urban management~\cite{chen2020spatial, liu2018line}, global resources monitoring~\cite{kennedy2009remote}, and damage assessment~\cite{bovolo2007split, zheng2021building}.

Early conventional CD methods mainly rely on hand-crafted features~\cite{ru2020multi}, which significantly limit the detection accuracy for high-resolution and very high-resolution remote sensing CD due to their low robustness and insufficient semantics. Recently, with the development of deep Convolutional Neural Networks (CNNs), deep learning has made significant progress in many computer vision tasks, such as semantic segmentation~\cite{jin2021mining, zheng2020foreground}, object detection~\cite{cui2021tf, zheng2020hynet}, and image retrieval~\cite{kim2020proxy}. Motivated by the powerful feature representation capability of CNNs, many deep learning-based CD approaches have been proposed and have achieved superior performance compared with traditional CD methods~\cite{shafique2022deep}.

Although previous research has made extensive efforts to improve the accuracy of CD results~\cite{shafique2022deep}. For example, Zhou et al.~\cite{zhou2022spatial} utilized the multi-head self-attention mechanism to capture the inter-temporal information and reduce the imaging difference so that the changes can be detected easily. Wang et al.~\cite{wang2023cross} introduced two attentive feature aggregation schemes to merge the cross-layer features in different processes. Fang et al.~\cite{fang2023changer} proposed a feature interaction module that consists of a set of alternative interaction layers in the feature extractor to perform temporal feature interaction. In some real-world applications, it is crucial to determine the changed regions completely, while previous methods fail to detect changes in the hard regions since they pay equal attention to all changed and unchanged regions. In recent years, the hard samples mining strategies have been extensively explored in various fields, e.g., data clustering~\cite{wang2023cross}, image segmentation~\cite{gu2020hard}, multi-modal learning~\cite{hu2021learning}. Liu et al.~\cite{wang2023cross} designed a sample-weighted strategy to lead the network mine not only the hard negative samples but also the hard positive sample. With the hard sample mining manner, the network can capture more discriminative features for the data clustering task. Gu et al.~\cite{gu2020hard} employs two depth-related metrics, i.e., depth prediction error and depth-aware segmentation error to evaluate the hard pixels, from which a weight map is formulated to encourage the model to pay more attention to the hard pixels. Hu et al.~\cite{hu2021learning} presented two robust loss functions to make the deep networks focus on clean samples instead of noisy ones, thereby alleviating the interference of noisy samples. Motivated by the great success of hard sample mining strategies in various research areas, we propose to integrate the hard samples mining manner into the change detection task so that the model can provide accurate change results for some real-world applications.

Based on the observations discussed above, we propose a novel change detection network, called HRANet, which is designed to generate accurate change maps under the guidance of a hard region mining manner. Firstly, we design a cross-layer knowledge review module that effectively captures the significant information of multi-level temporal difference features and avoids dilution of knowledge in a U-shape network. The cross-layer knowledge review module employs two attention mechanisms, namely reverse attention and conflict attention, to review the knowledge among multi-level temporal difference features. The reverse attention enables the network to refocus on the regions that are not segmented in the high-level features branch, while the conflict attention forces the network to pay more attention to the conflict regions between detection results from high-level and low-level features. This way, knowledge can be distilled from low-level into high-level features, thereby improving the feature presentation capability of temporal difference features. Secondly, we introduce an online hard region estimation branch, which dynamically estimates pixel-wise hard samples, supervised by the errors between predicted change maps and corresponding ground truth during training. Finally, the comprehensive knowledge of hard region aware features captured from the online hard region estimation branch and multi-level temporal difference features are integrated for accurate change detection. 

The main contributions of this work are summarized as follows:
\begin{enumerate}
	\item We propose a novel change detection network, HRANet, which incorporates a hard region mining strategy into the change detection task to achieve more accurate change results.
	\item We introduce an online hard region estimation branch that generates pixel-wise hard region maps related to both changed and unchanged areas, supervised by the errors between predicted change maps and ground truth during training. The hard region features are subsequently integrated with multi-level temporal difference features to guide the attention of model towards difficult changes.
	\item We present a cross-layer knowledge review module that employs reverse attention and conflict attention to effectively distill temporal change knowledge from low-level features into high-level ones. This allows for a comprehensive exploration of multi-level temporal difference features to identify changes more effectively.
	\item We conduct a series of experiments on benchmark datasets to evaluate the effectiveness and superiority of our proposed HRANet. The experimental results demonstrate that our approach outperforms various state-of-the-art models across two benchmark datasets.
\end{enumerate}

This paper is organized as follows. Section~\ref{sec:sec2} provides a brief review of related work in change detection. Section~\ref{sec:sec3} presents the proposed method, including the network architecture, online hard region estimation branch, and cross-layer knowledge review module. In Section~\ref{sec:sec4}, extensive experiments are conducted on two benchmark datasets to evaluate the performance of the proposed HRANet, with detailed discussions and model analyses. Finally, Section~\ref{sec:sec5} concludes this paper.

%-------------------------------------------------------------------------
\section{Related Work}\label{sec:sec2}
\subsection{Traditional CD Methods}
Traditional change detection (CD) methods primarily employ image transformation~\cite{celik2009unsupervised, crist1985tm} and image algebra~\cite{bruzzone2000automatic, coppin1996digital, sun2022structured, sun2021sparse,sun2021structure} techniques to identify changes in low- or middle-resolution remote sensing images. Previous methods typically utilize principal component analysis~\cite{celik2009unsupervised} and tasseled cap transformation~\cite{crist1985tm} to enhance change information in bi-temporal images, thereby enabling easy detection of changes. Representative image algebra techniques-based methods, such as image difference~\cite{bruzzone2000automatic}, image regression~\cite{coppin1996digital, sun2022structured, sun2021sparse,sun2021structure}, and change vector analysis~\cite{lambin1994change} are used to formulate difference images and choose suitable thresholds to identify changes. However, traditional CD methods rely on hand-crafted features that lack sufficient contextual information, resulting in limited performance for high-resolution and very high-resolution remote sensing CD.

\subsection{Deep Learning-based CD Methods}
Deep Learning-based CD Methods, with their powerful feature representation capability, have become the dominant solutions for high-resolution and very high-resolution remote sensing CD recently~\cite{fang2021snunet,ru2020multi,zheng2021change,zhan2017chang, zhang2018triplet,liu2016deep,wang2020deep, zhao2019incorporating, tang2021unsupervised}. Previous methods boost the accuracy of CD mainly from the following aspects. 

1) Multi-level feature fusion: As we know, the features extracted from the top layers of backbones are beneficial to locating objects, while those extracted from the bottom layers of backbones are conducive to recovering the details of objects~\cite{lin2017feature}. In recent years, many works with various well-designed multi-level feature fusion architectures are proposed to boost CD performance~\cite{shafique2022deep,peng2019end,fang2021snunet,ru2020multi,zheng2021change,zhan2017chang, zhang2018triplet,liu2016deep,wang2020deep, zhao2019incorporating, tang2021unsupervised}. For example, the long short-term memory (LSTM) and skip connection are combined to obtain more distinguishable multi-level features~\cite{papadomanolaki2021deep}. A full-scale feature fusion manner is introduced to aggregate multi-level information at diverse feature scales. Afterward, hybrid attention is utilized to capture the long-range context information among bi-temporal features for a more accurate CD. Li et al.~\cite{Li2022cd} proposed a guided refinement model, which first aggregates multi-level features and then exploits the aggregated features to repetitively polish multi-level features, to filter out the irrelevant noise information mixed in multi-level features. More inspiring related works can refer to the recent survey \cite{shafique2022deep}.

2) Temporal difference extraction: Effectively capturing the temporal difference information from bi-temporal features also plays a key component in CD. Previous methods usually employ feature concatenation or subtraction operations to realize temporal difference extraction. As demonstrated in~\cite{zheng2022changemask}, the original bi-temporal information can be preserved by the temporal concatenation operation, which stacks bi-temporal features along the dimension of channels in order. The feature subtraction needs an additional hand-crafted distance, e.g., $\ell_1$ or $\ell_2$ distance, to measure the pixel-wise temporal difference. Apart from the above two native temporal difference extraction manners, Zheng et al.~\cite{zheng2022changemask} considered the temporal order and designed a temporal-symmetric transformer, which exploits weight-shard 3-D convolutional layers with different temporal arrange orders to extract temporal difference information. Lei et al.~\cite{lei2021difference} introduced a difference enhance module (DEM) based on the feature subtraction measure to select the most significant channels of bi-temporal features with channel attention. Li et al.~\cite{Li2022cd} considered the complementary aspects of feature concatenation or subtraction operation and designed a temporal feature interaction module, in which the change regions of bi-temporal features can be highlighted.

3) Attention mechanisms: In recent years, attention mechanisms have received intensive attention, due to their abilities in capturing the context information and boosting feature discriminability. Channel attention~\cite{hu2018squeeze}, convolution block attention module~\cite{woo2018cbam}, and non-local~\cite{wang2018non} are three widely used attention mechanisms in the recently proposed CD method. 

Previous methods have made significant efforts to improve the accuracy of CD from multi-level feature fusion, temporal difference extraction, and attention mechanisms, etc. In various real-world applications, accurately identifying changed areas is essential. Previous methods often fall short in detecting changes in challenging regions, as they treat all areas in bi-temporal images with equal importance. This motivates us to propose a novel CD method with the hard region mining strategy to force the model to pay more attention to the challenging regions.

\subsection{Hard Sample Mining Strategy}
The purpose of hard sample mining is to enhance the discriminative capability of the deep networks. In recent years, the hard sample mining strategy has progressively become an active research topic in some deep learning-based tasks, such as face recognition~\cite{schroff2015facenet, xiao2022ihem}, object detection~\cite{oksuz2020imbalance}, person re-identification~\cite{chen2020hard}. The hard sample mining strategy focuses on selecting samples that are challenging to classify correctly, such as hard negatives and hard positives. This approach helps alleviate the imbalance between positive and negative samples, ultimately enhancing model training. For example, Nie et al. ~\cite{nie2019difficulty} introduced a sample selection strategy based on estimated confidence, encouraging the network to focus more on hard samples. A pixel-adaptive convolution filter is employed to estimate confidence, enhance the upsampling operation, and effectively propagate the high-confidence pixels into less reliable regions ~\cite{wannenwetsch2020probabilistic}. Liu et al. ~\cite{wang2023cross} developed a sample-weighted strategy that enables the network to mine not only hard negative samples but also hard positive samples. By employing this hard sample mining approach, the network can capture more discriminative features for data clustering tasks. Inspired by the significant success of hard sample mining strategies across various research domains, we propose incorporating this approach into the change detection task, to enhance the ability of the model to deliver accurate change results for real-world applications. Previous hard sample mining methods typically use score maps to identify hard samples, which are then applied as weights in the loss function to encourage the model to focus on these challenging samples. In contrast, we directly aggregate hard region-aware features with multi-level temporal difference features, enabling the model to better recognize changes in difficult regions.

%------------------------------------------------------------------------
\section{Proposed Method}\label{sec:sec3}
In this section, the overview of the proposed HRANet is first presented. Next, the structures of the online hard region estimation manner and cross-layer knowledge review module are given in detail. Finally, we illustrate the loss function of HRANet.

\subsection{Overview}
\begin{figure*}[!htbp]
	\centering
	\includegraphics[width=0.9\textwidth]{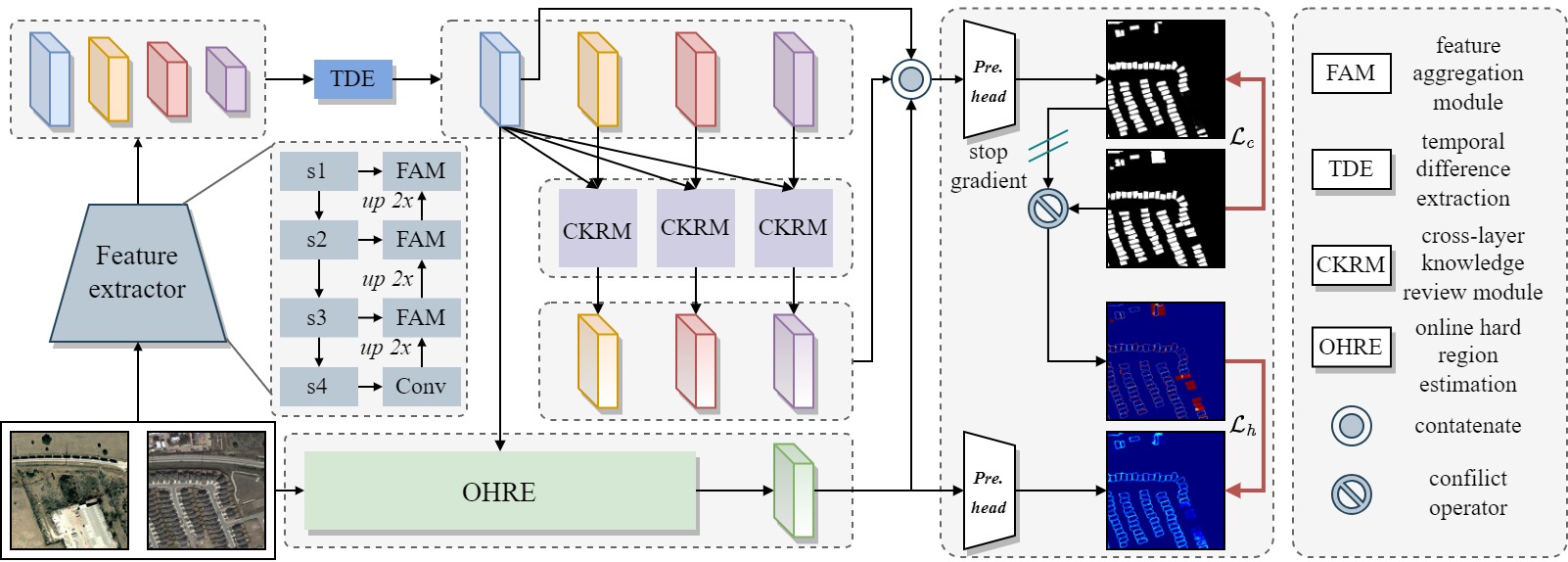}
	\caption{A framework of the proposed HRANet. 
		Initially, the bi-temporal images pass through a shared feature extractor to obtain bi-temporal features, and then multi-level temporal difference features are obtained through the TDE. The OHRE branch estimates pixel-wise hard samples corresponding of changed and unchanged regions, supervised by the diversity between predicted change maps and corresponding ground truth in the training process. CKRMs fully explore the multi-level temporal difference knowledge to enhance the feature capabilities. Finally, the multi-level temporal difference features and hard region-aware features obtained from the OHRE branch are aggregated to generate the final change maps.}
	\label{fig:HRANet}
\end{figure*}

\begin{figure*}[!htbp]
	\centering
	\includegraphics[width=0.9\textwidth]{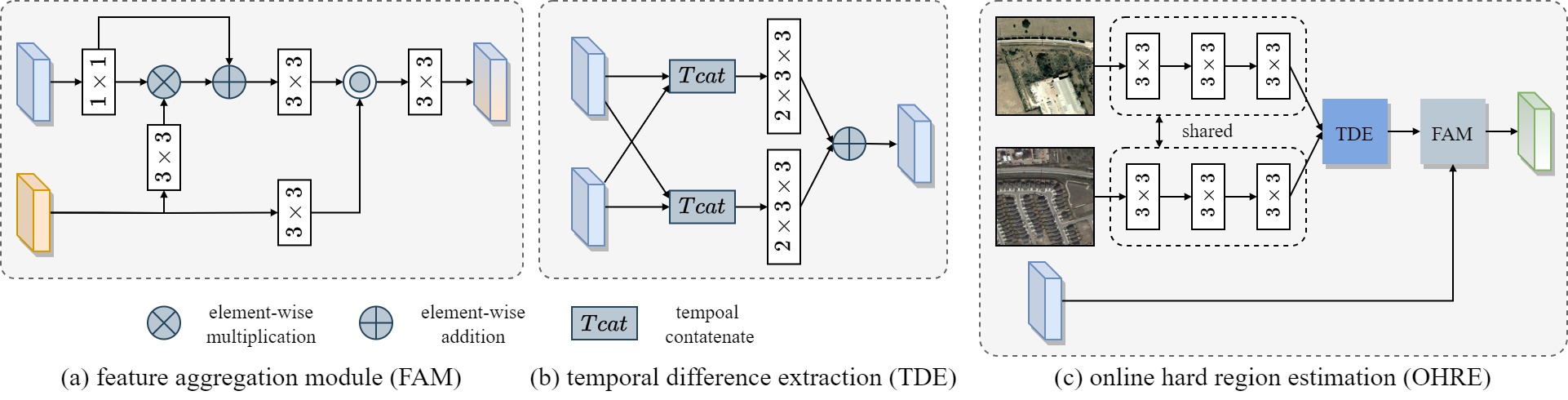}
	\caption{Illustration of the feature aggregation module (FAM), temporal difference extracting (TDE), and online hard region estimation (OHRE) branch.}
	\label{fig:OHRE}
\end{figure*}

As shown in Fig.~\ref{fig:HRANet}, we first utilize a shared feature extractor, which follows an encoder-decoder structure, to extract bi-temporal features. For the encoder, any backbone can be used here, and we utilize a ResNet18~\cite{he2016deep} as done in previous methods~\cite{chen2020spatial, chen2021remote, Li2022cd} for a fair comparison. Let a registered pair of images be denoted as $\mathbf{I}^t \in \mathbb{R}^{h\times w \times 3}, t\in\{1, 2\}$, where $h$, $w$, and $t$ denote the height, width and temporal order of images, respectively. The multi-level bi-temporal features, denoted as $\mathbf{F}_i^t \in \mathbb{R}^{h/2^{i+1} \times h/2^{i+1} \times c_i}, i\in\{2, 3, 4, 5\}, t\in\{1, 2\}$, can be obtained from the four stages of the encoder $\mathrm{Enc}(\cdot)$, as,
\begin{equation}
	\mathbf{F}_i^t = \mathrm{Enc}(\mathbf{I}^t), i\in\{2, 3, 4, 5\}, t\in\{1, 2\} 
\end{equation}
then, the decoder is implemented in a U-shape structure, in which a feature aggregation module (FAM) is introduced to progressively fuse the multi-level context information from high-levels to low-levels. As shown in Fig.~\ref{fig:OHRE} (a), FAM first computes an element-wise weight from the high-level feature to calibrate the low-level feature. Then high-level and low-level features are concatenated and pass through a convolution layer for feature fusion. With the decoder, we capture multi-level bi-temporal features $\mathbf{P}_i^t \in \mathbb{R}^{h/2^{i+1} \times h/2^{i+1} \times c_i}, i\in\{2, 3, 4, 5\}, t\in\{1, 2\}$ for detecting changes, as,
\begin{equation}
	\mathbf{P}_i^t =
	\begin{cases} 
		\mathbf{F}_i^t,  & \mbox{if } i=5\\
		\mathrm{FAM}(\mathbf{F}_i^t, \mathbf{F}_i^{t-1}), & \mbox{if } i\in\{2, 3, 4\}
	\end{cases}
\end{equation}
where $\mathrm{FAM}(\cdot)$ denotes the FAM. As discussed in the previous section, the temporal difference extraction (TDE) is a key component of CD. Here, we adopt the temporal-symmetric transformer proposed in~\cite{zheng2022changemask} (as shown in Fig.~\ref{fig:OHRE} (b)) to achieve TDE. The bi-temporal features are concatenated with two diverse temporal orders and pass through 3D convolution layers to capture temporal difference information. The procedure of TDE can be represented as,
\begin{equation}
	\mathbf{D}_i = \mathcal{T}(\mathbf{P}_i^1, \mathbf{P}_i^2), i\in\{2, 3, 4, 5\},
\end{equation}
where $\mathcal{T}(\cdot)$ represents the function of TDE, $\mathbf{D}_i$ denotes the obtained temporal difference features.


As illustrated previously, the U-shape structure progressively aggregates the multi-level information layer by layer. However, the long dependence between high-level and low-level features and the discriminative changed regions are ignored in such feature fusion strategy, resulting dilution of distinctive knowledge. To this end, the proposed knowledge review technique is exploited to fully capture the complementary information among multi-level temporal difference features for a more accurate CD. On the other hand, an online hard region estimation branch is designed to evaluate pixel-wise hard regions to strength the capability of model to deal with the hard detected changes in bi-temporal images. From the online hard region estimation branch, the hard region-aware features are further combined with multi-level temporal difference features to make the network pay more attention to the less reliable regions. More details of the knowledge review technique and online uncertainty estimation will be presented in Subsection \ref{sub:KR} and \ref{sub:OHRE}, respectively.

\subsection{Online Hard Region Estimation}\label{sub:OHRE}
The goal of hard region estimation is to determine the confidence related to the prediction of model on a pixel-wise basis. To achieve this, we introduce an online hard region estimation (OHRE) branch, as shown in Figure~\ref{fig:OHRE} (c). OHRE first uses three down-convolution blocks, implemented with three $3\times 3$ convolutional layers, followed by batch normalization and ReLU functions, to extract texture information from the bi-temporal images. The temporal difference extracting (TDE) method is then employed to capture temporal difference information, which is further combined with the temporal difference features $\mathbf{D}_2$ via a FAM to capture comprehensive knowledge for uncertainty estimation. The entire process can be expressed as follows,
\begin{equation}
	\begin{split}
		&\mathbf{T}^t = \mathrm{DoB}(\mathbf{I}^t), t\in\{1,2\}, \\
		&\mathbf{U} = \mathcal{F}(\mathbf{D}_2, \mathcal{T}(\mathbf{T}^1, \mathbf{T}^2)),
	\end{split}
\end{equation}
where $\mathrm{DoB}(\cdot)$ and $\mathcal{F}(\cdot)$ denote the down-convolution blocks and FAM, respectively. $\mathbf{T}^t$ and $\mathbf{U}$ represent the bi-temporal texture features and uncertainty-aware features, respectively.

Intuitively, the network should have higher confidence in the true positive and false negative regions of the prediction but lower confidence in the false positive and false negative regions. Therefore, we aim to use the differences between the prediction of model and the corresponding ground truth as the supervision signal to guide the learning process of the OHRE branch. The hard regions supervision can be formulated as follows,
\begin{equation}
	\mathbf{g}^u = \mathbf{p}^c \cdot (\mathbf{1} - \mathbf{g}^c) + \mathbf{g}^c \cdot (\mathbf{1} - \mathbf{p}^c),
\end{equation}
where $\mathbf{g}^u$, $\mathbf{p}^c$, and $\mathbf{g}^c$ represent the hard regions supervision, predicted change map, and corresponding true changes, respectively. We train the OHRE branch with a binary cross-entropy (BCE) loss $\mathcal{L}_{bce}$ which is formulated as,
\begin{equation}
	\mathcal{L}_u = \mathcal{L}_{bce}(\mathbf{p}^u, \mathbf{g}^u) = \mathbf{g}^u \cdot \mathrm{log}(\mathbf{p}^u) + (\mathbf{1} - \mathbf{g}^u) \cdot \mathrm{log}(\mathbf{1} -\mathbf{p}^u),
\end{equation} 
where $\mathcal{L}_u$ denotes the hard region loss, $\mathbf{p}_u$ is the predicted hard region map.

\subsection{Cross-layer Knowledge Review Module}\label{sub:KR}

\begin{figure}[!htbp]
	\centering
	\includegraphics[width=0.48\textwidth]{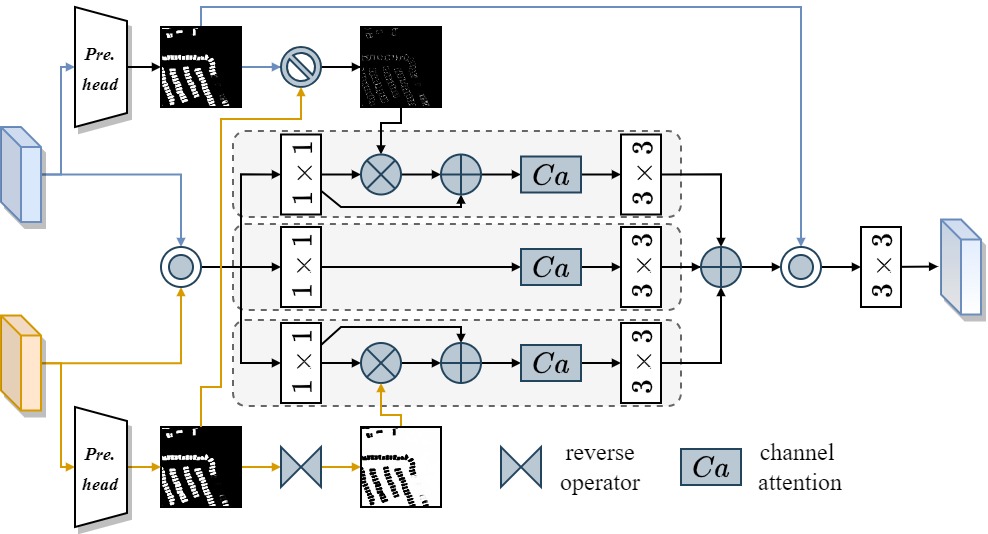}
	\caption{Illustration of the cross-layer knowledge review module (CKRM).}
	\label{fig:CKRM}
\end{figure}

It is commonly acknowledged that the low-level features extracted from the bottom layers of the backbone contain abundant object details, while the high-level features obtained from the top layers of the backbone consist of semantic abstractions about the objects. To achieve accurate CD, it is crucial to fully utilize the information in both low-level and high-level features. Although the U-shaped structure of the feature extractor performs multi-level feature fusion, it fails to adequately explore the long dependence between high-level and low-level features and the discriminative changed regions, leading to the dilution of distinctive knowledge. To address this issue, we propose a knowledge review strategy consisting of three cross-layer knowledge review modules (CKRM). As illustrated in Fig.~\ref{fig:CKRM}, a multi-branch structure equipped with conflict attention and reverse attention is utilized to capture the complementary information between the temporal difference features $\mathbf{D}_2$ and $\mathbf{D}_3$.

As mentioned earlier, the multi-level temporal difference features capture diverse aspects of changed objects. To force the network to focus more on the conflicts between changed regions of high-level and low-level features and to extract the complementary information, we implement conflict attention. Let $\mathbf{p}_2^c$ and $\mathbf{p}_3^c$ denote the predicted change maps of temporal difference features $\mathbf{D}_2$ and $\mathbf{D}_3$, the conflicts regions can be calculated as the difference between $\mathbf{p}_2^c$ and $\mathbf{p}_3^c$,
\begin{equation}
	\mathrm{CoA} = \mathbf{p}_2^c \cdot (\mathbf{1} - \mathbf{p}_3^c) + \mathbf{p}_3^c \cdot (\mathbf{1} - \mathbf{p}_2^c),
\end{equation}
where $\mathrm{CoA}$ is the obtained conflict attention between $\mathbf{D}_2$ and $\mathbf{D}_3$.

In addition to conflict attention, we also introduce reverse attention to seek complementary information across multiple levels. High-level features are effective for object localization, but provide less detail about changed objects. Therefore, we exploit the low-level features to compensate for the insufficient detail in high-level features, guided by the reverse attention. Specifically, the reverse attention is formulated to erase the changed regions predicted by high-level features, thereby forcing the network to focus more on the unchanged regions. The reverse attention is formulated as,
\begin{equation}
	\mathrm{ReA} = (\mathbf{1} - \mathbf{p}_3^c),
\end{equation}
where $\mathrm{ReA}$ denotes the reverse attention.

Afterwords, CKRM concatenates $\mathbf{D}_2$ and $\mathbf{D}_3$ and pass through three $1\times 1$ convolution layers for feature transition as,
\begin{equation}
	\begin{split}
		&\mathbf{D}_1^k = \mathrm{Conv}_{1\times 1}(\mathrm{Cat}(\mathbf{D}_2, \mathbf{D}_3)),\\
		&\mathbf{D}_2^k = \mathrm{Conv}_{1\times 1}(\mathrm{Cat}(\mathbf{D}_2, \mathbf{D}_3)),\\
		&\mathbf{D}_3^k = \mathrm{Conv}_{1\times 1}(\mathrm{Cat}(\mathbf{D}_2, \mathbf{D}_3)),
	\end{split}
\end{equation}
where $\mathbf{D}_1^k$, $\mathbf{D}_2^k$, and $\mathbf{D}_3^k$ denote the features for three branches of CKRM. $\mathrm{Cat}(\cdot)$ is the feature concatenate operator. The conflict attention and reverse attention are plugged into two branches making the network capture the discriminative changed regions. In addition, a channel attention~\cite{hu2018squeeze} subsequent with a $3\times 3$ convolution layer is also injected into each branch of CKRM for enhancing the feature representation capability. The whole process can be denoted as,
\begin{equation}
	\begin{split}
		&\mathbf{D}_1^r = \mathrm{Conv}_{3\times 3}(\mathrm{CA}(\mathbf{D}_1^k \oplus \mathbf{D}_1^k\otimes \mathrm{CoA})),\\
		&\mathbf{D}_2^r = \mathrm{Conv}_{3\times 3}(\mathrm{CA}(\mathbf{D}_2^k \oplus \mathbf{D}_2^k\otimes \mathrm{ReA}),\\
		&\mathbf{D}_3^r = \mathrm{Conv}_{3\times 3}(\mathrm{CA}(\mathbf{D}_3^k)),
	\end{split}
\end{equation}
where $\mathbf{D}_1^r$, $\mathbf{D}_2^r$, $\mathbf{D}_3^r$ are the enhanced features. $\mathrm{CA}(\cdot)$, $\oplus$, and $\otimes$ represent the channel attention, element-wise addition, and multiplication, respectively. Finally, $\mathbf{D}_1^r$, $\mathbf{D}_2^r$, $\mathbf{D}_3^r$ and integrated into one representation, and further concatenate with the prediction $\mathbf{p}_{2c}$ to generate final temporal difference features $\hat{\mathbf{D}}_3$ as,
\begin{equation}
	\hat{\mathbf{D}}_3 = \mathrm{Conv}_{3\times 3}(\mathrm{Cat}(\mathbf{p}_{2c}, \mathbf{D}_1^r\oplus \mathbf{D}_2^r\oplus \mathbf{D}_3^r))
\end{equation}

As shown in Fig.~\ref{fig:HRANet}, three CKRM are inserted in HRANet to distill the fine-gained temporal change information into high-level features so as to fully capture the complementary information among multi-level temporal difference features. In this way, our HRANet can provide more accurate change maps.

\subsection{Loss Function}
We jointly train the change detection part and online hard region estimation branch. For change detection part, a hybrid loss~\cite{Li2022cd, li2023lightweight} including a BCE loss $\mathcal{L}_{bce}$ and a dice (Dice) loss $\mathcal{L}_{dice}$ \cite{milletari2016v}, is adopted as,
\begin{equation}
	\mathcal{L}_c = \mathcal{L}_{bce} + \mathcal{L}_{dice}
\end{equation}
where $\mathcal{L}_c$ is the change detection loss.

With the hard region supervision, the total training loss of HRANet can be formulated as,
\begin{equation}
	\mathcal{L} = \mathcal{L}_{c} + \mathcal{L}_{u}
\end{equation}

\section{Experiments}\label{sec:sec4}

\begin{table*}[!htbp]
	\centering
	\caption{Quantitative comparisons of the proposed method with diverse settings in terms of $\kappa$, IoU, F1, OA, Rec, and Pre on LEVIR+ and BCDD datasets.}
	\label{tab:ab}
	\resizebox{0.9\textwidth}{!}{%
		\begin{tabular}{@{}cccccccccccc@{}}
			\toprule[1.5pt]
			\multicolumn{1}{c|}{\multirow{2}{*}{No.}} & \multicolumn{1}{c|}{\multirow{2}{*}{Variants}} & \multicolumn{5}{c|}{LEVIR + (in-domain testing)}                                      & \multicolumn{5}{c}{BCDD (out-domain testing)}                     \\ \cmidrule(l){3-12} 
			\multicolumn{1}{c|}{}                     & \multicolumn{1}{c|}{}                          & $\kappa$ & IoU    & F1     & Rec    & \multicolumn{1}{c|}{Pre}    & $\kappa$ & IoU    & F1     & Rec    & Pre    \\ \midrule
			\multicolumn{1}{c|}{\#1}                  & \multicolumn{1}{c|}{HRANet}                  & 0.8582   & 0.7605 & 0.8640 & 0.8618 & \multicolumn{1}{c|}{0.8662} & 0.7683   & 0.6370 & 0.7783 & 0.7656 & 0.7913 \\ \midrule
			\multicolumn{12}{l}{(a) Feature Aggregation Module (FAM)}                                                                                                                                                     \\ \midrule
			\multicolumn{1}{c|}{\#2}                  & \multicolumn{1}{c|}{FAM w/o Gate}              & 0.8503   & 0.7488 & 0.8564 & 0.8458 & \multicolumn{1}{c|}{0.8671} & 0.7445   & 0.6059 & 0.7546 & 0.6880 & 0.8353 \\ \midrule
			\multicolumn{12}{l}{(b) Online Hard Region Estimation (OHRE)}                                                                                                                                                  \\ \midrule
			\multicolumn{1}{c|}{\#3}                  & \multicolumn{1}{c|}{w/o OHRE}                   & 0.8541   & 0.7543 & 0.8600 & 0.8485 & \multicolumn{1}{c|}{0.8717} & 0.7558   & 0.6203 & 0.7656 & 0.7114 & 0.8288 \\ \midrule
			\multicolumn{1}{c|}{\#4}                  & \multicolumn{1}{c|}{OHRE w/o HRL}               & 0.8556   & 0.7565 & 0.8614 & 0.8507 & \multicolumn{1}{c|}{0.8724} & 0.7437   & 0.6043 & 0.7534 & 0.6672 & 0.8651 \\ \midrule
			\multicolumn{1}{c|}{\#5}                  & \multicolumn{1}{c|}{OHRE w B sup} & 0.8444   & 0.7402 & 0.8507 & 0.8431 & \multicolumn{1}{c|}{0.8585} & 0.7487   & 0.6112 & 0.7587 & 0.6969 & 0.8325 \\ \midrule
			\multicolumn{12}{l}{(c) Cross-layer Knowledge Review Module (CKRM)}                                                                                                                                                        \\ \midrule
			\multicolumn{1}{c|}{\#6}                  & \multicolumn{1}{c|}{w/o CKRM}                   & 0.8209   & 0.7065 & 0.8280 & 0.8148 & \multicolumn{1}{c|}{0.8417} & 0.7367   & 0.5960 & 0.7469 & 0.6749 & 0.8361 \\ \midrule
			\multicolumn{1}{c|}{\#7}                  & \multicolumn{1}{c|}{CKRM w/o CoA}                & 0.8467   & 0.7436 & 0.8530 & 0.8579 & \multicolumn{1}{c|}{0.8481} & 0.7771   & 0.6476 & 0.7861 & 0.7355 & 0.8442 \\ \midrule
			\multicolumn{1}{c|}{\#8}                  & \multicolumn{1}{c|}{CKRM w/o ReA}                & 0.8478   & 0.7452 & 0.8540 & 0.8526 & \multicolumn{1}{c|}{0.8553} & 0.7671   & 0.6352 & 0.7769 & 0.7491 & 0.8069 \\ \midrule
			\multicolumn{1}{c|}{\#9}                  & \multicolumn{1}{c|}{CKRM w/o ReA \& CoA}          & 0.8399   & 0.7336 & 0.8463 & 0.8384 & \multicolumn{1}{c|}{0.8544} & 0.7474   & 0.6096 & 0.7575 & 0.6968 & 0.8297 \\ \bottomrule[1.5pt]
		\end{tabular}%
	}
\end{table*}

\subsection{Datasets}
In our experiments, we use two remote sensing change detection benchmark datasets to evaluate the effectiveness of the proposed HRANet. The detailed information is listed as follows,

\textbf{LEVIR+}~\cite{shen2021s2looking}: It is a expanded version of LEVIR~\cite{chen2020spatial}, which consists of 637 pairs of bi-temporal remote sensing images captured from the Google Earth platform with $1024\times 1024$ spatial size and $0.5$m spatial resolution. LEVIR+ includes an additional 348 bi-temporal remote sensing images, with both datasets mainly focusing on building changes. In our experiments, we merge the validation and testing set of LEVIR to form a training set for LEVIR+. The additional 348 bi-temporal remote sensing images are used as a separate testing set. Due to limited GPU memory, all images are cropped into non-overlapping $512\times 512$ patches. A total of 1780/768/1392 image pairs are obtained for training/validation/testing, respectively.

\textbf{BCDD} \footnote{http://study.rsgis.whu.edu.cn/pages/download/building\_dataset.html}: It is a building change detection dataset containing a $32507\times 15354$ bi-temporal images with $0.075$m resolution. In our experiments, we none overlap crop the dataset into $512\times 512$ size of patches and then obtained a total of 1827 pairs of bi-temporal images for the out-domain testing.

\subsection{Evaluation Metrics}
To verify the effectiveness of the proposed HRANet, five widely used evaluation metrics~\cite{ding2022bi, lei2021difference, huang2021multiple}, namely Kappa coefficient ($\kappa$), intersection over union (IoU), F1-score (F1), recall (Rec), and precision (Pre) are adopted in our experiments. The detailed calculations of five metrics can be found in ~\cite{Li2022cd, li2023lightweight}.

\subsection{Implementation details}
We implement the proposed HRANet via the Pytorch toolbox \cite{paszke2019pytorch}. Random flipping, cropping, and temporal exchanging are employed on the image patches for data augmentation. The AdamW~\cite{loshchilovdecoupled} optimizer is exploited to optimize the proposed HRANet, setting the weight decay and parameters $\beta_1$, $\beta_2$ to 0.01, 0.9, and 0.99, respectively. We adopt the $poly$ learning scheme to adjust the learning rate as $(1 - \frac{cur\_iteration}{max\_iteration} )^{power} \times lr$, setting $power$ and $max\_iteration$ to 0.9 and 20000, respectively. The batch size is 8 and the initial learning rate is set as 0.0005. A single Nvidia Titan V GPU is used for training, validation, and testing.

\subsection{Ablation Studies}
We evaluate the effectiveness of the components and configurations of the proposed HRANet on LEVIR+ and BCDD datasets and the diverse change detection performance measured by $\kappa$, IoU, F1, Rec, and Pre in Tab.~\ref{tab:ab}.

\subsubsection{Effectiveness of feature aggregation module (FAM)}
As demonstrated in the previous section, the Feature Aggregation Module (FAM) utilizes an element-wise weight calculated from high-level features to calibrate low-level features for better feature aggregation. To validate the effectiveness of FAM, we constructed an approach without the element-wise weight scheme and denoted it as FAM w/o G (\#02 in TABLE \ref{tab:ab}). The results in Tab.~\ref{tab:ab} show that when the element-wise weight scheme is removed, the change detection performance drops significantly. This indicates the effectiveness of the proposed FAM and emphasizes the importance of the element-wise weight scheme in achieving optimal performance.

\subsubsection{Effectiveness of online hard region estimation (OHRE)}
In our proposed HRANet, OHRE is a key component for formulating the pixel-wise confidence related to the predicted change results and performing hard region learning. To verify its effectiveness, we first remove OHRE module to set a w/o OHRE method (\#03 in TABLE \ref{tab:ab}), of which the change detection performance is reported in TABLE \ref{tab:ab}. As seen from the results, the change detection performance decreases respectively about 1.25, 1.67, and 1.23 percentages in terms of $\kappa$, IoU, and F1 on BCDD datasets, indicating the necessity of OHRE module. In addition, to demonstrating the effectiveness of hard region learning in OHRE, we only exploit the multi-level temporal difference features without the features from the OHRE branch to obtain the final change detection, which is terms as OHRE w/o HRL (\#05 in TABLE \ref{tab:ab}). We can observe that the hard region learning manner can sightly improve the change detection performance, verifying its effectiveness. 

To achieve OHRE, we constructed a pseudo hard region supervision by modeling the difference between ground truth change maps and the final predicted change maps. To verify the effectiveness of the pseudo hard region supervision, we replaced it with the change boundary supervision and referred to this method as OHRE w B sup (\#04 in TABLE \ref{tab:ab}). From the results in TABLE \ref{tab:ab}, we observed that the pseudo hard region supervision obtained the best performance. This verify the effectiveness of the proposed pseudo hard region supervision.

\subsubsection{Effectiveness of Cross-layer knowledge review module (CKRM)}
We apply the CKRM to distill the fine temporal change knowledge into high-level features and jointly use the multi-level temporal difference feature to generate the final change maps. The CKRM consists of two key attention mechanisms: reverse attention and conflict attention. To validate the effectiveness of CKRM, we formulate four variants of the proposed HRANet, denoted as w/o CKRM, CKRM w/o CoA, CKRM w/o ReA, and CKRM w/o CoA \& ReA, as \#06-\#09 in TABLE \ref{tab:ab}. In w/o CKRM, we remove the CKRM and only use the initial multi-level temporal difference features and uncertainty-aware features to obtain the final change maps. For CKRM w/o CoA and CKRM w/o ReA, we separately remove the conflict attention and reverse attention. For CKRM w/o CoA \& ReA, we remove both the conflict attention and reverse attention from the network. From the results in TABLE \ref{tab:ab}, we observe that using only the reverse attention or conflict attention has worse performance than the default setting. Therefore, the default setting of CKRM, which uses both reverse attention and conflict attention, is the most effective choice.

\begin{table*}[!htbp]
	\centering
	\caption{Quantitative comparisons in terms of $\kappa$, IoU, F1, OA, Rec, and Pre on LEVIR+ and BCDD datasets. The best and second best results are highlighted in {\color[HTML]{FF0000} red} and {\color[HTML]{00B0F0} blue}, respectively.}
	\label{tab:main_res}
	\resizebox{0.9\textwidth}{!}{%
		\begin{tabular}{@{}cc|ccccccccccc@{}}
			\toprule[1.5pt]
			\multicolumn{2}{c|}{Datasets}                       & FC-diff & FC-ef  & FC-cat & L-Unet & DSIFN                        & SNUNet & BIT    & MSCANet                      & TFI-GR                       & A2Net  & Ours                         \\ \midrule
			\multicolumn{2}{c|}{FPS}                            & 124.30  & 178.31 & 122.76 & 28.93  & 15.60                        & 14.42  & 44.85  & 53.78                        & 77.23                        & 56.48  & 28.85                        \\ \midrule
			\multicolumn{2}{c|}{FLOPs (G)}                      & 37.66   & 28.50  & 42.49  & 138.52 & 658.22                       & 438.47 & 84.46  & 117.03                       & 82.81                        & 24.08  & 112.78                       \\ \midrule
			\multicolumn{2}{c|}{Params (M)}                     & 1.35    & 1.35   & 1.54   & 8.45   & 50.71                        & 12.04  & 3.50   & 17.11                        & 27.88                        & 3.78   & 14.34                        \\ \midrule
			\multicolumn{1}{c|}{\multirow{5}{*}{LEVIR+ (in-domain testing)}} & $\kappa$ & 0.7148  & 0.5242 & 0.6346 & 0.7820 & {\color[HTML]{00B0F0}0.8386} & 0.7578 & 0.8175 & 0.8278                       & 0.8074                       & 0.8252 & {\color[HTML]{FF0000}0.8582} \\
			\multicolumn{1}{c|}{}                       & IoU   & 0.5705  & 0.3738 & 0.4822 & 0.6541 & {\color[HTML]{00B0F0}0.7318} & 0.6230 & 0.7016 & 0.7161                       & 0.6884                       & 0.7125 & {\color[HTML]{FF0000}0.7605} \\
			\multicolumn{1}{c|}{}                       & F1    & 0.7266  & 0.5442 & 0.6507 & 0.7909 & {\color[HTML]{00B0F0}0.8451} & 0.7677 & 0.8246 & 0.8346                       & 0.8154                       & 0.8321 & {\color[HTML]{FF0000}0.8640} \\
			\multicolumn{1}{c|}{}                       & Rec     & 0.7349  & 0.5617 & 0.7080 & 0.7918 & 0.8324                       & 0.7736 & 0.7954 & 0.8124                       & {\color[HTML]{00B0F0}0.8345} & 0.8127 & {\color[HTML]{FF0000}0.8618} \\
			\multicolumn{1}{c|}{}                       & Pre     & 0.7184  & 0.5277 & 0.6019 & 0.7899 & {\color[HTML]{00B0F0}0.8582} & 0.7619 & 0.8561 & 0.8580                       & 0.7972                       & 0.8525 & {\color[HTML]{FF0000}0.8662} \\ \midrule
			\multicolumn{1}{c|}{\multirow{5}{*}{BCDD (out-domain testing)}}  & $\kappa$ & 0.6103  & 0.4863 & 0.5037 & 0.6951 & 0.7312                       & 0.6284 & 0.7019 & 0.6473                       & {\color[HTML]{00B0F0}0.7449} & 0.7433 & {\color[HTML]{FF0000}0.7683} \\
			\multicolumn{1}{c|}{}                       & IoU   & 0.4544  & 0.3351 & 0.3507 & 0.5478 & 0.5896                       & 0.4745 & 0.5555 & 0.4924                       & {\color[HTML]{00B0F0}0.6078} & 0.6051 & {\color[HTML]{FF0000}0.6370} \\
			\multicolumn{1}{c|}{}                       & F1    & 0.6249  & 0.5020 & 0.5193 & 0.7079 & 0.7418                       & 0.6436 & 0.7142 & 0.6599                       & {\color[HTML]{00B0F0}0.7561} & 0.7539 & {\color[HTML]{FF0000}0.7783} \\
			\multicolumn{1}{c|}{}                       & Rec     & 0.5466  & 0.3866 & 0.4068 & 0.6816 & 0.6777                       & 0.6065 & 0.6784 & 0.5590                       & {\color[HTML]{00B0F0}0.7603} & 0.7217 & {\color[HTML]{FF0000}0.7656} \\
			\multicolumn{1}{c|}{}                       & Pre     & 0.7292  & 0.7158 & 0.7180 & 0.7363 & {\color[HTML]{FF0000}0.8193} & 0.6856 & 0.7541 & {\color[HTML]{00B0F0}0.8052} & 0.7520                       & 0.7892 & 0.7913                       \\ \bottomrule[1.5pt]
		\end{tabular}%
	}
\end{table*}

\begin{figure*}[!htbp]
	\centering
	\includegraphics[width=1\textwidth]{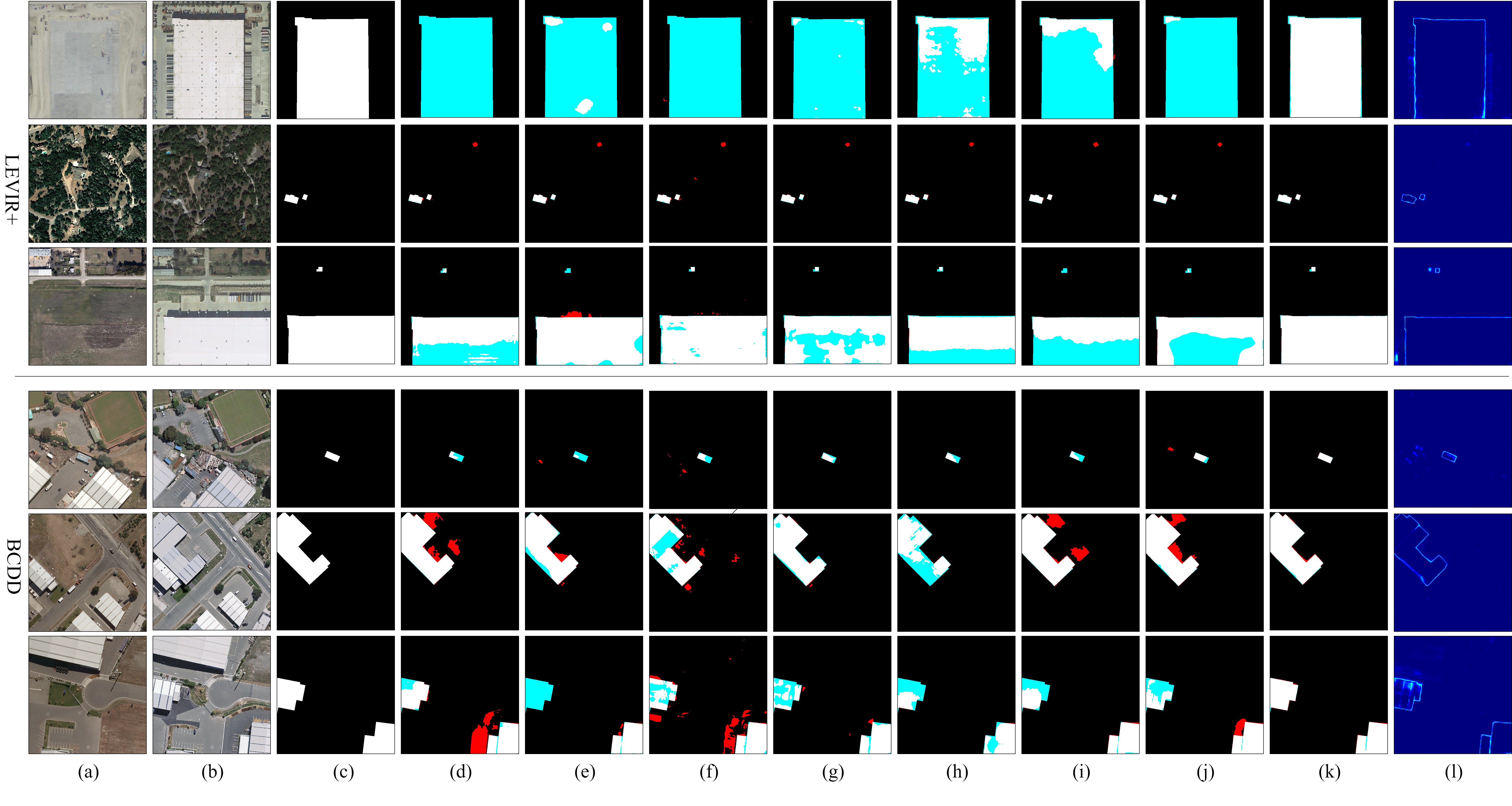}
	\caption{Visual comparisons of the proposed method and the state-of-the-art approaches on the LEVIR+ dataset. (a) $t_1$ images; (b) $t_2$ images; (c) Ground-truth; (d) L-Unet; (e) DSIFN; (f) SNUNet; (g) BIT; (h) MSCANet; (i) TFI-GR; (j) A2Net; (k) Ours, (l) Predicted hard region maps. The rendered colors represent true positives (white), false positives ({\color[HTML]{FF0000} red}), true negatives (black),  and false negatives ({\color[HTML]{00B0F0} blue}).}
	\label{fig:vis_CD}
\end{figure*}

\subsection{Comparison With the State-of-the-art}
We compare the proposed model with ten state-of-the-art remote sensing change detection approaches, including 
FC-Diff~\cite{daudt2018fully}, 
FC-ef~\cite{daudt2018fully}, 
FC-Cat~\cite{daudt2018fully},
L-Unet~\cite{papadomanolaki2021deep}, 
DSIFN~\cite{zhang2020deeply}, 
SNUNet~\cite{fang2021snunet}, 
BIT~\cite{chen2021remote},
MSCANet~\cite{Liu_2022_CD},
TFI-GR~\cite{Li2022cd},
and A2Net~\cite{li2023lightweight}. 
We reproduce the change detection results by using their released code under their default parameters for fair comparisons.

\subsubsection{Quantitative evaluation}\label{sec:qe1}
We present the quantitative evaluation results of various methods for remote sensing change detection, measured by $\kappa$, IoU, F1, Rec, and Pre, in Table~\ref{tab:main_res}. In addition, Table~\ref{tab:main_res} provides the FPS (Frames Per Second), model parameters (Params), and computation costs (FLOPs). The results show that the proposed HRANet outperforms other methods in both in-domain and out-domain testing. For instance, on the LEVIR+ dataset (in-domain testing), the proposed HRANet achieves approximately 1.96, 2.87, and 1.89 percentage higher performance improvements than the second-best method (DSIFN). On the BCDD dataset (out-domain testing), the proposed HRANet achieves around 2.34, 2.92, and 2.22 percentage higher performance improvements than the second-best method (TFI-GR).  Additionally, the proposed HRANet has comparable FPS, model parameters, and computation costs to some heavy models, such as L-Unet, SNUNet, and DSIFN. These findings strongly demonstrate the effectiveness and superiority of the proposed HRANet.

\subsubsection{Qualitative evaluation}\label{sec:qe2}
We present visual comparisons of various methods on LEVIR+ and BCDD datasets in Fig.~\ref{fig:vis_CD}. White, red, black, and blue are used to indicate true positives, false positives, true negatives, and false negatives, respectively, for better visualization. Based on the visual results, we observe that the proposed method demonstrates superiority in the following aspects:
\begin{enumerate}
	\item \textit{Advantages in Content Integrity}: 
	In Fig.~\ref{fig:vis_CD}, it can be observed that some of the compared methods, such as L-Unet, DSIFN, SNUNet, BIT, MSCANet, TFI-GR, and A2Net, fail to detect the entire changed regions. On the other hand, the proposed HRANet can accurately identify the changed objects with fine content integrity. The superior performance of our approach is attributed to two main reasons. Firstly, we employ a knowledge review approach to distill the fine-grained change information into coarse layers, and we jointly utilize the multi-level change information to comprehensively generate accurate change results. Secondly, our method estimates the hard regions, and we use a hard region learning approach to encourage the network to focus more on such difficult regions, resulting in more precise change detection results.
	\item \textit{Advantages in Location Accuracy}: 
	The visual results in Fig.~\ref{fig:vis_CD} demonstrate that the proposed HRANet accurately identifies the changed regions and effectively eliminates the irrelevant pseudo changes caused by background clutters in bi-temporal images. In contrast, several other methods, such as L-Unet, DSIFN, SNUNet, BIT, MSCANet, TFI-GR, and A2Net, fail to eliminate the irrelevant pseudo changes. These results underscore the efficacy of the proposed HRANet.
	\item \textit{Advantages in Detecting changes in hard regions}: 
	The proposed HRANet incorporates an hard region estimation approach to provide pixel-wise confidence levels associated with the predicted change maps. Our results in Fig.~\ref{fig:vis_CD} indicate that the estimated hard region maps are mainly concentrated along the boundary of the changed objects, indicating high confidence when the predicted change maps are accurate. This highlights the effectiveness of the proposed hard region estimation approach.
\end{enumerate}

\section{Conclusion}\label{sec:sec5}
In this work, we present a novel change detection network designed to deliver accurate results in challenging regions. Our proposed method designs a cross-layer knowledge review module and an online hard region estimation branch, both of which enhance change detection performance, particularly in difficult cases. Experimental results from two high spatial resolution remote sensing change detection datasets show that our approach surpasses existing state-of-the-art methods in this domain.

%% use section* for acknowledgment
\ifCLASSOPTIONcompsoc
\section*{Acknowledgments}
\else
\section*{Acknowledgment}
\fi

The authors wish to gratefully acknowledge the anonymous reviewers for the constructive comments of this paper.

\ifCLASSOPTIONcaptionsoff
  \newpage
\fi

\bibliographystyle{IEEEtran}
\bibliography{IEEEabrv,References}

\end{document}